\documentclass[10pt, conference]{IEEEtran}
\IEEEoverridecommandlockouts

\usepackage[bookmarks=false]{hyperref}
\usepackage{amsmath}
\usepackage{enumitem,kantlipsum}
\usepackage{graphicx}
\usepackage[table]{xcolor}
\usepackage{booktabs}
\usepackage{threeparttable}
\usepackage{tabulary}
\usepackage{multirow}
\usepackage{nicefrac}
\usepackage{fancyhdr}
\usepackage[binary-units=true]{siunitx}

\usepackage[compact]{titlesec}
\titlespacing{\section}{0pt}{*1}{*0}
\titlespacing{\subsection}{0pt}{*1}{*0}
\titlespacing{\subsubsection}{0pt}{*0}{*0}
\usepackage[left=0.5in,right=0.5in,top=0.5in,bottom=1in,nofoot]{geometry}

\newcommand{\revISPLED}[1]{{\color{black}#1}}
\newcommand{\revLUCA}[1]{{\color{black}#1}}
\newcommand{\revDAC}[1]{{\color{black}#1}}

\IEEEaftertitletext{\vspace{-1.2\baselineskip}}

\IEEEoverridecommandlockouts
\begin{document}
\bstctlcite{IEEEexample:BSTcontrol}
\title{TCN Mapping Optimization for Ultra-Low Power Time-Series Edge Inference}

\author{\IEEEauthorblockN{Alessio Burrello\IEEEauthorrefmark{1}, Alberto Dequino\IEEEauthorrefmark{1}, Daniele Jahier Pagliari\IEEEauthorrefmark{2}, Francesco Conti\IEEEauthorrefmark{1}, \\Marcello Zanghieri\IEEEauthorrefmark{1}, Enrico Macii\IEEEauthorrefmark{2}, Luca Benini\IEEEauthorrefmark{1}\IEEEauthorrefmark{2}, Massimo Poncino\IEEEauthorrefmark{2}}

\IEEEauthorblockA{\IEEEauthorrefmark{1}University of Bologna, Bologna, Italy. \textit{Email: name.surname@unibo.it}\\
\IEEEauthorrefmark{2}Politecnico di Torino, Turin, Italy. \textit{Email: name.surname@polito.it}}
}

\maketitle
\begin{abstract}
Temporal Convolutional Networks (TCNs) are emerging lightweight Deep Learning models for Time Series analysis. We introduce an automated exploration approach and a library of optimized kernels to map TCNs on Parallel Ultra-Low Power (PULP) microcontrollers. Our approach minimizes latency and energy by exploiting a layer tiling optimizer to jointly find the tiling dimensions and select among alternative implementations of the causal and dilated 1D-convolution operations at the core of TCNs.
We benchmark our approach on a commercial PULP device, achieving up to \revDAC{103$\times$ lower latency and 20.3$\times$} lower energy than the Cube-AI toolkit executed on the STM32L4 and from 2.9$\times$ to 26.6$\times$ lower energy compared to commercial closed-source and academic open-source approaches on the same hardware target.
\end{abstract}
\begin{IEEEkeywords}
Internet-of-Things, Deep-Learning, Edge-Computing, Temporal Convolutional Network.

\end{IEEEkeywords}
\section{Introduction}
\label{sec:intro}
\thispagestyle{fancy}
\fancyhf{}
\chead{Published as a conference paper at the 2021 IEEE/ACM International Symposium on Low Power Electronics and Design (ISLPED)}
The Internet of Things (IoT) envisions billions of interrelated computing devices, favoring the growth of a broad spectrum of applications, such as personalized healthcare~\cite{zanghieri2019robust}, structural health monitoring and robotics~\cite{duisterhof2019learning}.
To deal with the increasing amount of data produced by IoT systems, a promising recent trend consists of performing part of the computation directly within the sensing device. This relieves the network from some pressure and results in faster and more predictable response times, challenging to achieve with cloud-based systems.
Moreover, it also positively impacts battery life and privacy by eliminating the need to transmit large amounts of (possibly sensitive) raw data through energy-hungry wireless links~\cite{flamand2018gap}. 

Time-series processing using Deep Learning (DL) models is at the core of many of the aforementioned IoT applications~\cite{zanghieri2019robust,duisterhof2019learning}.
Therefore, researchers are putting a significant effort into developing new methodologies and tools to enable the execution of these computationally intensive models at the edge~\cite{ shangguan2019optimizing}. 
Temporal Convolutional Networks (TCNs)~\cite{bai2018empirical} are a sub-class of one-dimensional Convolutional Neural Networks (1D-CNNs) that include features specifically tailored for time series.
%
Recently, impressive results in terms of accuracy and energy efficiency have been obtained, applying TCNs to tasks such as biosignal analysis~\cite{zanghieri2019robust}, and predictive maintenance~\cite{ren2020cloud}.

In this work, we propose an efficient and parallelized implementation of the main  kernels involved in TCN inference, explicitly tailored for memory-constrained edge devices. 
To the best of our knowledge, ours is the first work targeting \revISPLED{the minimization of latency and energy} of general \revDAC{1D-CNN and} TCN topologies at the edge.
Our main contributions are as follows:
\begin{enumerate}[leftmargin=*,topsep=0pt]
\item We introduce a new software library that implements multiple alternative strategies to run the main TCN computational kernels, tailored for optimal execution on PULP devices.
\item We build accurate execution time models of our kernels, which \revISPLED{are used to automatically map the layers to the most efficient kernel implementation based on their geometrical parameters.}
\item We demonstrate the effectiveness of our multi-kernel library and selection strategy by comparing our work with state-of-the-art NN backends \revDAC{targeting 1D-CNNs and TCNs}. 
With experiments on the 8-core cluster of the GAP8 processor~\cite{flamand2018gap}, \revDAC{we obtain up to 17.2 MAC/cycles, with an average performance improvement on single layers of $9.7\times$ with respect to the proprietary backend for the same chip and up to $354\times$ compared to the Cube-AI toolkit on a STM32H7 microcontroller (MCU).} %
\item We \revDAC{show that our toolkit enables end-to-end real-world TCNs to run on extreme edge devices by deploying} three complete state-of-the-art TCNs~\cite{zanghieri2019robust,bai2018empirical}, achieving up to 1.11 GMAC/s and an energy efficiency of 21.79 GMAC/s/W.
\end{enumerate}

\revDAC{Compared to the state-of-the-art, we achieve up to $103\times$ lower latency and $20.3\times$ lower energy on full networks.}
On the gesture recognition application of \cite{zanghieri2019robust}, this allows us to meet real-time constraints with an end-to-end latency of 13.6ms while \revISPLED{reducing the energy consumption from 0.9 mJ per inference of the original paper to 0.69 mJ \revLUCA{at a power consumption of 51 mW}. }
\vspace{-0.1cm}

\section{Background and Related Works}
\label{sec:background}

\subsection{Temporal Convolutional Networks}
\label{subsec:tcn} 
%
The fundamental building blocks of TCNs are 1D convolutional layers~\cite{bai2018empirical,lea2016temporal}.
With respect to standard 1D-CNNs, however, TCN layers include two fundamental elements: \emph{causality}, which implies that each convolution output $\mathbf{y}_{t}$ is only a function of inputs $\mathbf{x}_{\tilde{t}}$ with $\tilde{t} \leq t$ and \emph{dilation}, a fixed step $d$ introduced between inputs to increase the receptive field of the layer on the time axis without requiring more parameters.
Thus, a convolutional layer in a TCN implements the following function:
\vspace{-0.3cm}
\begin{equation}\label{eq:1d_conv}
\mathbf{y}_t^m = \text{Conv}\,(\mathbf{x}) = \sum_{i=0}^{K-1} \sum_{l=0}^{C_{in}-1} \mathbf{x}_{t-d\,i}^l \cdot \mathbf{W}_i^{l,m}
\vspace{-0.2cm}
\end{equation}
where $\mathbf{x}$ is the input feature map, $\mathbf{y}$ the output feature map, $t$ the time index, $\mathbf{W}$ the filter weights, $C_{in}$ the number of input channels, $m$ the output channel, $d$ the dilation factor, and $K$ the filter size.
In the original paper~\cite{bai2018empirical}, TCNs were proposed as fully-convolutional architectures. Modern embodiments also include other common layers such as pooling and linear ones~\cite{ren2020cloud, zanghieri2019robust}, which are analogous to those present in 1D-CNNs. However, as for standard CNNs, most of the computational complexity of TCNs comes from convolutional layers~\cite{zanghieri2019robust}; therefore, the peculiarities of these kernels have a fundamental impact on inference optimization.
\subsection{IoT end-nodes}\label{sec:iot}
Modern IoT end-nodes are increasingly based on Parallel Ultra-Low-Power (PULP) Systems-on-Chip (SoC), often composed of a single control MCU coupled with one or more Digital Signal Processing (DSP)-oriented processors.
For instance, STM \cite{st-dual} and NXP \cite{nxp} have recently introduced dual-core SoCs combining an ARM Cortex-M0 and an ARM Cortex-M4, where the former mainly deals with I/O and the latter has DSP-specific functionalities such as single-cycle Multiply-and-ACcumulate (MAC) instructions and Single-Instruction Multiple Data (SIMD) capabilities.
Similarly, GreenWaves Technologies' GAP8 SoC~\cite{flamand2018gap} features one I/O core and an 8-core cluster with a RISC-V Instruction Set Architecture (ISA) extension for enhanced DSP.
With the introduction of these more advanced computing engines, also the memory hierarchy shows an increased complexity.
All these end-nodes include at least a small, but fast L1 memory for fast accesses and a wider L2 for data storage.
A data management infrastructure is usually also included to explicitly manage these memories, allowing for the creation of hand-crafted software caching mechanism.
For example, GAP-8 includes a general-purpose Direct Memory Access (DMA) controller to move data between memories, reducing the memory access bottleneck.

\subsection{Software backends for DNN inference at the edge}
Although custom accelerators enable the execution of NN inference at the edge with the highest efficiency, these dedicated hardware devices are only economically affordable in high-end systems~\cite{TPU}. Therefore, academia and industry have started to investigate software backends that maximize NN inference efficiency on  \textit{general purpose} edge devices such as those mentioned in Sec.~\ref{sec:iot}~\cite{garofalo2020pulp, lai2018cmsis,wang2020fann,GWTAutotiler}, leveraging their DSP-oriented capabilities~\cite{gautschi2017near}.
To cope with the tight memory constraints of edge devices, these libraries focus on \textit{quantized} kernel implementations, in which weights ($\mathbf{W}$) and feature maps ($\mathbf{x}$/$\mathbf{y}$) are stored using low-precision integers, typically on 8-bits~\cite{lai2018cmsis}.

CMSIS-NN~\cite{lai2018cmsis} and PULP-NN~\cite{garofalo2020pulp} are among the most relevant state-of-the-art open-source libraries for the execution of quantized kernels on ARM and RISC-V cores, respectively.
Both focus on standard CNN kernels and leverage a smart data organization (both in terms of static layout and input reordering at runtime) to convert 2D convolutions into Matrix-Matrix multiplications (MatMul), which are then processed with SIMD operations to maximize performance.
Moreover, both libraries optimize the MatMul phase by partially unrolling the inner loop so that \revDAC{16--32} MACs are computed in each iteration (e.g., using the \texttt{sdotp} vector operation in~\cite{garofalo2020pulp}), and data reuse is maximized.
There are also other backends for NN inference at the edge, but they either focus on a limited set of kernels, e.g. handling only fully-connected layers \cite{wang2020fann}, or are closed source~\cite{GWTAutotiler}. %

On top of these libraries, full-stack solutions that automatically deploy a pre-trained neural network to a target device have been proposed, such as Cube-AI~\cite{CubeAI} from STMicroelectronics, GWT NN-Tool~\cite{GWTAutotiler}, or the open-source DORY~\cite{burrello2020dory}.
These tools ingest a NN dataflow description from popular training frameworks (e.g. TensorFlow or PyTorch) and produce a complete C application for the target edge processor.
To deal with large layers that do not entirely fit the memory level closest to the core (L1), these tools exploit \textit{tiling} mechanisms to divide them into chunks with a lower memory footprint.
Besides memory occupation, optimal tiling parameters are determined also considering the corresponding backend's implementation details, e.g. avoiding tiles that would not permit to exploit unrolled/SIMD computations fully.

We can divide the aforementioned tools and libraries into two groups.
Some, e.g. PULP-NN and CMSIS-NN, do not support causality and dilation in their convolution kernels, i.e., the two fundamental properties of TCNs, whose effect can only be reproduced by inserting zeros in the filters weights, negatively affecting performance and memory.
Instead, the backends of GWT NN-Tool and Cube-AI claim to support dilated convolutions \revDAC{(although the latter does not support quantization for dilated layers)} but are not optimized to execute 1D kernels.
At the time of writing, we are not aware of any dedicated and optimized implementation of TCNs on memory-limited, low-power IoT end-nodes.

\section{TCN Kernel Toolkit}
\label{sec:TCN_toolkit}
\revDAC{In this section, 
we introduce the main design choices on which our TCN library is based, our kernel implementations and how they are plugged into an optimizer~\cite{burrello2020dory} for code generation.
\revDAC{
As a representative of single and multi-core IoT-oriented computing platform, we target the open-source RISC-V \textit{RV32IMC} ISA, extended with domain-specific extensions (\textit{XpulpV2}~\cite{gautschi2017near}) for efficient digital signal processing such as load/store with address post-increment and SIMD MAC down to 8-bit vector operands.}
\revDAC{Here, we focus on TCN-specific layers -- we implemented other layers (pooling, linear, etc.) similarly to~\cite{garofalo2020pulp}.}
}

\subsection{Design Choices}
\label{subsec:opt}
\begin{figure}[ht]
  \centering
\includegraphics[width=.8\columnwidth]{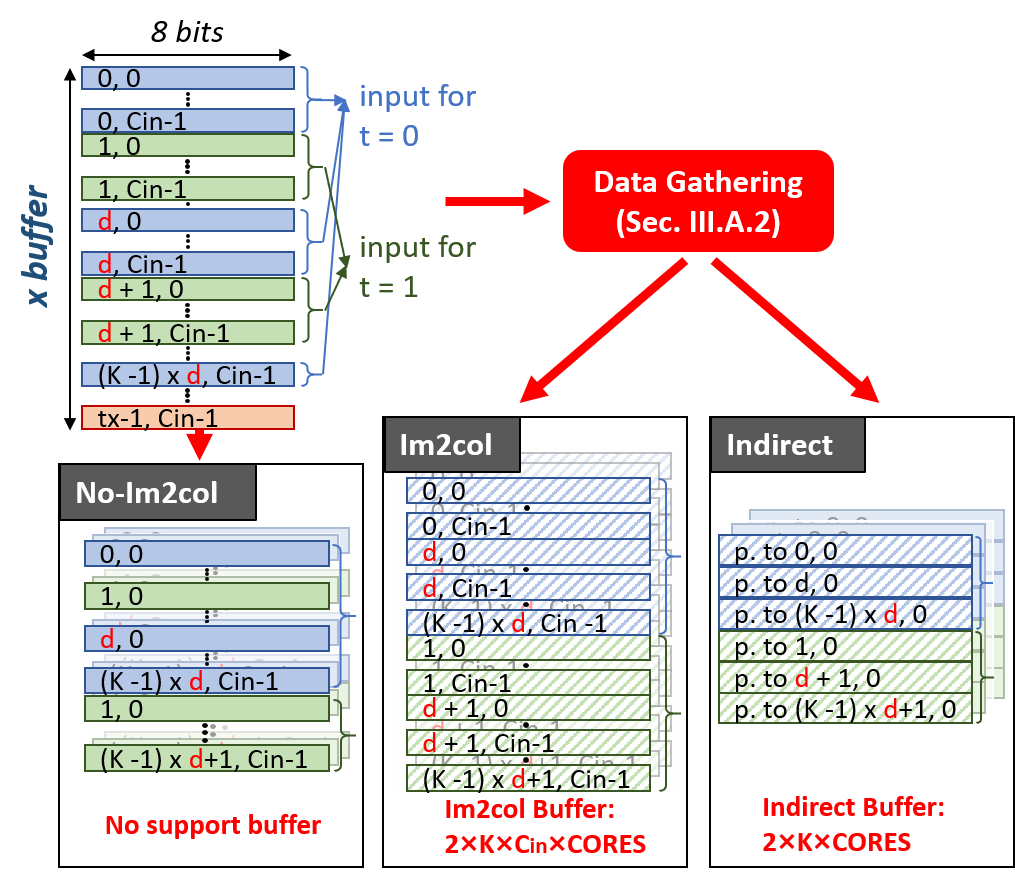}
  \vspace{-0.3cm}
  \caption{Three different input data gathering options used in the proposed kernels.}
  \label{fig:reordering}
  \vspace{-0.5cm}
\end{figure}

\textit{1) Data Layout:} Kernel libraries for 2D convolution organize input and output data either as Channel-Height-Width (CHW), i.e. with the spatial dimension as the innermost one, or  HWC. We call the equivalent layouts in the case of 1D convolutions CT (channel-time) and TC (time-channel).
We observe that using a TC layout,  the 1D convolution inputs of all channels of a single time-step (i.e. those required by the innermost summation of Eq.~\ref{eq:1d_conv}) are stored contiguously in memory. Inputs relative to subsequent time-steps are separated by $(d-1)\times C_{in}$ elements. In particular, for $d = 1$, all convolution inputs are stored contiguously.
Given the presence in many DSP-oriented ISAs of single-cycle loads with pointer increment (e.g., \texttt{p.lw} in \textit{XpulpV2}), we select the TC layout.
The chosen data organization is shown in the ``x buffer'' of Fig. \ref{fig:reordering}.

\textit{2) Data Gathering:} Conceptually, the convolution kernels in our library operate in two phases, which we call \textit{input data gathering} and \textit{MatMul loop} respectively, similarly to~\cite{garofalo2020pulp,lai2018cmsis}.
In the first phase, dedicated buffers in each core are used to prepare the input data needed for the convolution. Differently from CMSIS-NN and PULP-NN, we use both explicit \emph{im2col} buffers \cite{lai2018cmsis} and \emph{indirect} buffers \cite{dukhan2019indirect} in this phase.
%
%
\revDAC{This design choice is motivated by the fact that the two buffers exploit different trade-offs in terms of memory occupation and performance.
For instance, the indirect buffers allows to strongly reduce memory occupation on layers with a large $C_{in}$, allowing them to fit in small on-chip memories.}
\begin{figure}[ht]
  \centering
\includegraphics[width=.84\columnwidth]{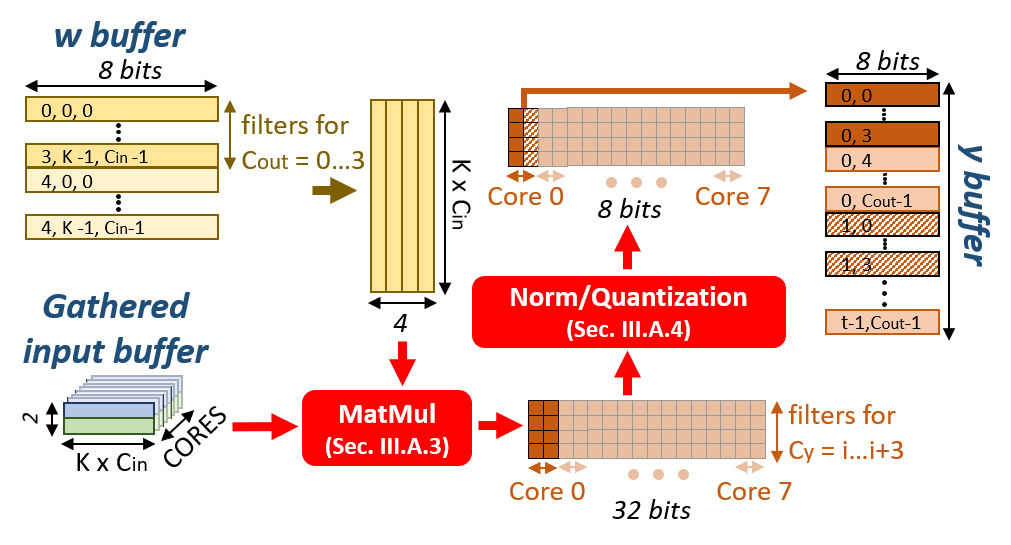}
  \vspace{-0.4cm}
  \caption{MatMul loop, Quantization and Batch Normalization in the proposed toolkit. Parallelization over multiple cores is represented by lighter colors.}
  \label{fig:matmul}
  \vspace{-0.6cm}
\end{figure}
Note that \revDAC{in non-dilated 1D-CNN convolutional layers ($d=1$)}, input data is already contiguous, and this phase can be bypassed.

\textit{3) MatMul Loop:} After data gathering, convolution reduces to a series of MatMul, as depicted in Fig. \ref{fig:matmul}.
We use a $4\times2$ unrolled MatMul (i.e. the product of 4 sets of weights with 2 sets of inputs) as our atomic operation. \revDAC{This is motivated by the result of~\cite{garofalo2020pulp}, where the authors found that 4$\times$2 unrolling maximizes data reuse in a RISC-V register file with 32 registers.}
Since 4$\times$2 unrolling requires two sets of inputs, we allocate two im2col/indirect buffers in each core (see Fig. \ref{fig:reordering}).
Each of the unrolled MatMuls is further vectorized using the \texttt{pv.sdotsp.b} instruction of the \textit{XpulpV2} ISA, which computes the dot product of 4 contiguously stored 8-bit inputs in parallel.

\textit{4) Normalization and Quantization}: We ``fuse'' the quantization and normalization pointwise operations, essential for quantized inference~\cite{burrello2020dory}, together with our convolution kernels. 
In contrast, using \textit{separate} kernels for these operations would result in additional data movement and worsen performance.

\textit{5) Parallelization:} We split the convolution workload on multiple cores over the time dimension, i.e., each core computes the output features of all channels for an assigned range of time-steps. 
We select time-wise over channel-wise parallelization since it allows cores to produce outputs of their assigned time-steps without exchanging partial data with other cores, and to store results on a separate, contiguous memory's area.
The workload subdivision among cores is shown on the right of Fig. \ref{fig:matmul}.

\subsection{1D Convolutional Kernels}
\label{subsec:kernels}
The three convolution kernels implemented in our library differ mainly in the data gathering phase, as shown in Figure \ref{fig:reordering}.

\textit{1) No-im2col Kernel:} As explained in Section \ref{subsec:opt}-1, due to the sequential nature of 1D data and the TC layout, data gathering can be bypassed when $d=1$.
Removing this buffering phase has a positive effect on both memory usage and performance.
On the other hand, for kernels with $d>1$, performing the MatMul loop without data gathering would require interleaving the weight vectors with zeros, to eliminate the contribution of input time-steps that have to be skipped. 
The resulting memory occupation increase and performance loss makes the \emph{No-im2col} approach feasible only for non-dilated convolutions, where, however, it is optimal for both memory and performance.

\textit{2) Im2col kernel:} To efficiently handle dilation rates higher than 1, data gathering becomes necessary. One approach is to use an im2col buffer~\cite{lai2018cmsis} (bottom-center of Figure~\ref{fig:reordering}). This is a linear array in which all inputs required to produce a given convolution output are copied contiguously (in Eq. 1, these are the $C_{in}$ input features relative to the $K$ $d$-spaced time-steps preceding step $t$). When the convolution stride is smaller than $K$, data will be replicated in multiple im2col buffers, causing a memory overhead. However, the linear im2col output yields maximal exploitation of the hardware facilities to optimize the MatMul performance (e.g. SIMD operations, single cycle pointer increment, etc.).
\revDAC{Notice that, by means of asynchronous DMA transfers, a double im2col buffering scheme could be used to mask the overhead of data gathering entirely. However, we discarded this approach since we found that, when combined with the tiling mechanism described in Sec.~\ref{sec:tiling}, the larger memory required by double buffering increases the number of tiles needed for a given layer, hence worsening the performance instead of improving it.}

%

%
\textit{3) Indirect kernel:} To minimize the memory footprint of convolution, the im2col buffer can be replaced with an \emph{indirect} buffer for data gathering. 
Instead of copying all convolution inputs in contiguous memory, this buffer only stores the \textit{pointers} to the first input relative to each time-step involved in the convolution (bottom-right of Fig.~\ref{fig:reordering}).
Indirect convolution reduces by a factor $C_{in}$ the memory overhead for data gathering but requires an additional loop to cycle through the buffer's addresses in the MatMul section, negatively impacting performance. To the best of our knowledge, ours is the first edge-oriented backend to include both im2col and indirect convolution kernels.
%

%
\subsection{Kernel modeling and selection}\label{sec:tiling}
\revDAC{
AI-oriented IoT endnodes often employ hierarchies of scratchpad memories instead of caches, to minimize power consumption and optimize data reuse patterns.
These devices often use small (32--128 KiB) memories with high-bandwidth at L1 and larger, lower-bandwidth ones at L2 and require explicit data tiling and DMA transfers to be used optimally.
In this work, we employed the DORY optimizer~\cite{burrello2020dory} as a baseline tool to automatically manage memory hierarchy, tiling layers in such a way that they can be executed over a small L1 and moving data between L1 and L2 with asynchronous DMA transfers.
}

\revDAC{
We modified the baseline optimizer so that not only it finds appropriate tiling solutions, but it also selects the optimal 1D convolution implementation for a given layer and tiling via an additional \emph{kernel selection} step.
%
Specifically, for convolutional layers, the optimizer first determines the best tiling scheme for each of the three alternative implementations using the Constraint Programming (CP) solver of~\cite{burrello2020dory}.
The tiling scheme is selected taking into account the different memory occupation of the \textit{no-im2col}, \textit{im2col} and \textit{indirect} kernel implementations.
Once tiling options are selected, the total execution cycles (\revISPLED{proportional to both latency and energy}) of the three layer implementations are extracted using a detailed model of the kernel, and the version with the highest performance is selected.

}

\begin{figure}
  \centering
\includegraphics[width=.77\columnwidth]{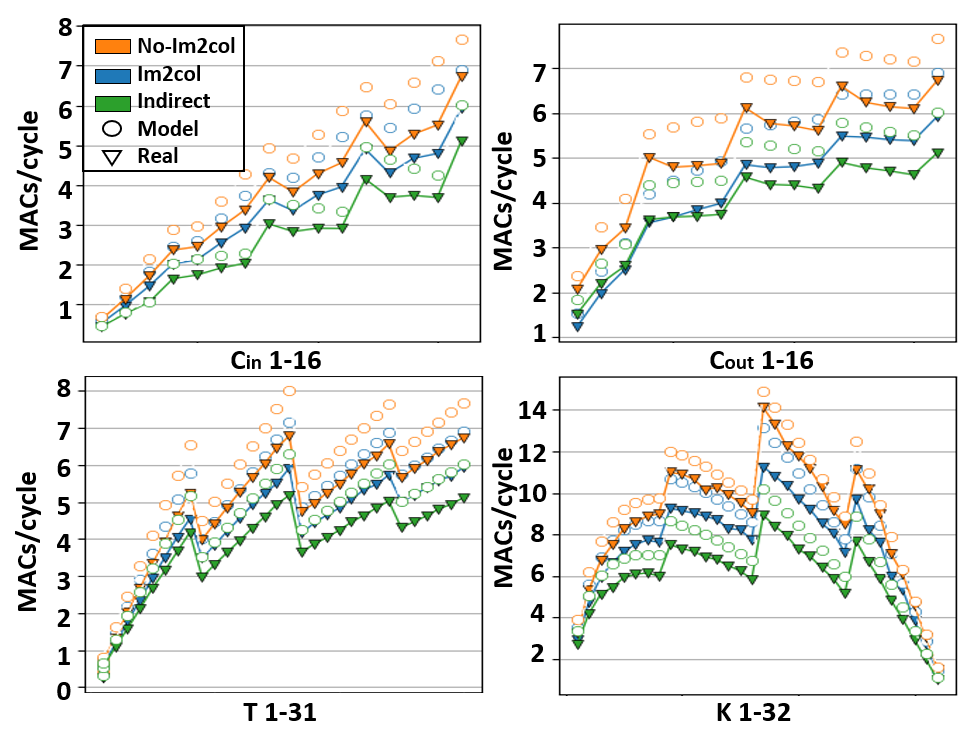}
  \vspace{-0.4cm}
  \caption{Modeling of our three kernels versus various layer parameters.}
  \label{fig:model}
  \vspace{-0.5cm}
\end{figure}
As an example, to model the performance of the \textit{im2col} kernel, we denote the total number of convolutions performed by each core as $\text{Core\_Iter} = \frac{T}{2N_{cores}}$, where $T$ is the total number of time-steps in the input sequence and the factor $2$ comes from the fact that all cores manage 2 time-steps simultaneously. We also call $\text{MM\_Iter} =  \frac{C_{out}}{4} $ the number of iterations on the output channel dimension performed within each convolution, where the factor 4 comes from the 4x2 MatMul loop that simultaneously generates 4 $C_{out}$ elements (Sec. \ref{subsec:opt}).

We then compute the execution cycles for the two main phases (data gathering and MatMul) and for the entire kernel as:
\begin{equation}\label{eq:gather}
\vspace{-0.5cm}
\text{Gather\_Cyc} =  \text{max (} 2 \times K \times \alpha, 2 \times K \times C_{in} \times \beta\text{)}
\end{equation}
\begin{equation}\label{eq:mm}
\vspace{-0.7cm}
\text{MM\_Cyc} = (\gamma + \delta \times C_{in} \times \nicefrac{K}{4} )
\end{equation}
\begin{equation}\label{eq:total}
\text{Cyc} = \text{Core\_Iter} \times (\epsilon + \text{Gather\_Cyc} + \text{MM\_Iter} \times \text{MM\_Cyc}).
\end{equation}
where $\alpha$, $\beta$, $\gamma$, $\delta$, $\epsilon$ are hardware-dependent constants corresponding to the cost in execution cycles for load/store, pointer updates, and arithmetic operations. 

The first equation derives from the use of asynchronous DMA transfers for data gathering. It computes the maximum between the DMA control overhead (first term, dependent on the $2\times K$ DMA invocations needed to build the two im2col buffers) and the cycles required for the actual transfer (second term, dependent on the size of the actual transmitted data). The MM\_Cyc equation computes the cycles of the MatMul loop, as a function of the layer parameters, where the division by 4 comes from the use of SIMD operations processing 4 8-bit elements per instruction. Here, the constants also account for batch normalization and quantization.

Models for the other two convolution kernels are built with similar reasoning and are not reported here for the sake of space. For the same reason, the reported equations do not consider corner cases (e.g. a number of time-steps not divisible by the number of cores), which are taken into account by the actual models.

Figure \ref{fig:model} shows the modeled vs. real performance of all kernels for different parameter sweeps.
Although there is an offset between real execution cycles and predicted ones, due to stalls and memory contentions, this gap is almost constant over all the parameters and kernels, hence not changing the ranking between different kernels' for a given set of parameters.

\begin{table}[t]
\footnotesize
\centering
\caption{Performance of our toolkit using different optimization criteria for tiling parameters and kernel selection.}
\vspace{-0.2cm}
\label{tab:kernel_model}
\begin{tabular}{c|c|c|c|c}
\multicolumn{1}{c|}{\multirow{2}{*}{\begin{tabular}[c|]{@{}c@{}}layer\\ ($C_{in}\times T \times C_{out}$)\end{tabular}}} & \multicolumn{3}{c|}{MACs/cycle}     \\ \cline{2-4} 
\multicolumn{1}{c|}{} & model  & heuristic & memory & best kernel\\ \hline
$64\times 256 \times 64$,  &&&&\\  $d = 1$, $K = 3$ & \textbf{15.98} & 15.47 & 12.50 & no-im2col\\ \hline
$256\times 16 \times 256$,   &&&&\\  $d = 2$, $K = 3$ & \textbf{14.92} & \textbf{14.92} & 4.14 & im2col\\ \hline
$1024\times 16 \times 1024$, &&&&\\ $d = 2$, $K = 3$ & \textbf{13.31} & 8.94 & 10.42 & indirect\\ \hline
\end{tabular}
\vspace{-0.4cm}
\end{table}
Table \ref{tab:kernel_model} shows the performance achieved combining our architectural models \revDAC{with the optimizer described at the beginning of this section. Results are reported for three different layer topologies.}
The table compares the results obtained with our cycles models with those obtained with other objective functions for the same tiling optimizer, namely the tiles' pure memory occupation and the simplified model used in~\cite{burrello2020dory} (column ``heuristic'').
As shown, our models achieve \revDAC{1.3$\times$/3.6$\times$} speed-up for complete layers with different geometries. This is mostly due to an accurate assessment of the execution time of border tiles, for which the computation loop might be under-utilized depending on the amount of data remaining to be computed.
Further, \revDAC{we note that all} three kernel variants \revDAC{are useful: each topology uses a different variant in its own optimal case}. \revISPLED{The column \emph{best kernel} shows the kernel which is selected by the tiling optimizer as the most efficient.} We confirmed by exhaustive search that the selected implementation is indeed the best for those parameters.

\section{Experimental Results and Discussion}
\label{sec:results}
We benchmark our toolkit on GAP-8~\cite{flamand2018gap}, a commercial PULP SoC including a control RISC-V processor (\textit{fabric controller}) and a \textit{cluster} of 8 additional RISC-V cores.
%
The cores within the cluster are 4-stage in-order single-issue pipelines called RI5CY~\cite{gautschi2017near}, \revDAC{which uses the \textit{RV32IMCXpulpV2} ISA mentioned in Section~\ref{sec:TCN_toolkit}}. The cluster's 8 cores share a 64 kB multi-banked L1 memory Tightly-Coupled Data Memory (TCDM), accessed through a high-bandwidth single-cycle-latency logarithmic interconnect. 
We compare our work with two state-of-the-art CNN backends for the same hardware target (PULP-NN~\cite{garofalo2020pulp} and GWT NN-Tool, on the GAP\_SDK v3.6~\cite{GWTAutotiler}), and with the Cube-AI toolchain (v5.1.2)~\cite{CubeAI} executed on the STM32H7, and the STM32L4 MCUs. 
\revDAC{All experiments refer to int8 quantized layers. Full networks are trained in a quantization-aware manner, with negligible accuracy loss compared to float versions.}
\revISPLED{We set GAP8, STM32H7, and STM32L4 frequencies at 100 MHz, 480 MHz, and 80 MHz, \revLUCA{with a corresponding power consumption of 51 mW, 234 mW, and 10 mW, respectively. We measured energies on the real devices.}
We use GMAC/s, GMAC/s/W, and MACs/cycle as comparison metrics. Note that while the first two are platform-dependent and thus most significant for back-ends on the same hardware (e.g., our toolkit, PULP-NN, and GWT NN-Tool), the latter is platform-independent and not linked to the frequency or power consumption of the specific platforms.}

\subsection{Kernels Comparison}
\label{subsec:kernel_res}
\begin{figure}
  \centering
\includegraphics[width=.77\columnwidth]{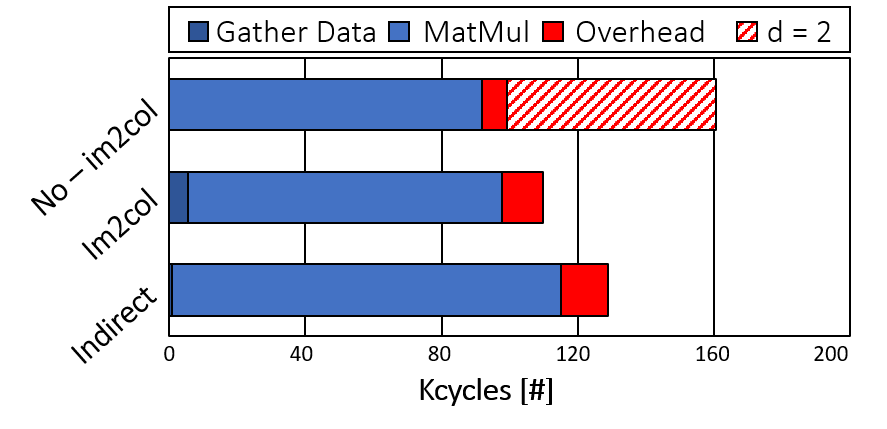}
  \vspace{-0.4cm}
  \caption{Execution cycles of our three 1D convolution kernels on a $64\times256\times32$ layer. The three kernels achieve 15.1, 13.7 and 12.0 MACs/cycle, respectively. With $d=2$, the No-im2col degradates to 9.7 MACs/cycle.}\label{fig:cycle}
  \vspace{-0.2cm}
\end{figure}

\begin{figure}
  \centering
\includegraphics[width=.8\columnwidth]{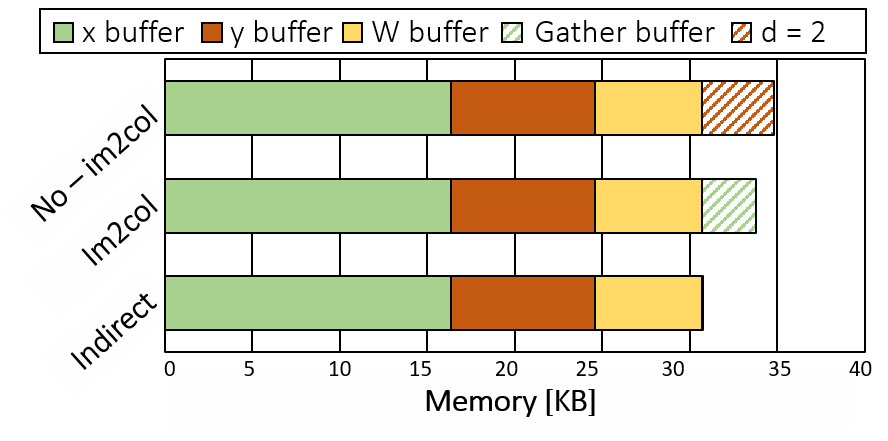}
  \vspace{-0.5cm}
  \caption{Memory occupation of the kernels of Fig.~\ref{fig:cycle}.}\label{fig:memory}
  \vspace{-0.6cm}
\end{figure}

Fig.~\ref{fig:cycle} and Fig.~\ref{fig:memory}  show a detailed analysis of our three 1D convolution implementations for a $64\times256\times32$ layer (i.e., $C_{in} = 64$, $T = 256$, $C_{out} = 32$) with $K = 3$.
Fig.~\ref{fig:cycle} reports the execution cycles for the data gathering and MatMul loop phases, and the additional cycles due to stalls and memory contentions, whereas Fig.~\ref{fig:memory} breaks down the memory occupation.
The graphs report the results for both $d=1$ and $d=2$, where the difference is only sizeable for the No-im2col kernel (diagonal red lines).
We select these layer parameters for the comparison so that the layer entirely fits into the 64KB L1 memory of GAP-8, and the effect of tiling does not alter the results.
Note that this structure could also be considered one tile of e.g., a wider $64\times256\times64$ layer.

For $d=1$, the No-im2col kernel obtains both the minimum number of cycles and the smallest memory occupation, since it does not require the creation of an additional gathering buffer (im2col or indirect). However, for $d > 1$, the same kernel has significant overheads in operations and memory due to the added zeros in the weight buffer. For $d = 2$, we see 62\% more operations and an additional 5 KB of memory, making the No-Im2col kernel the worst of the three. These overheads increase with larger $d$.

The Im2col kernel uses fewer instructions in the MatMul loop than the Indirect one while paying more time to create its gather buffer. In this example, the trade-off results in an overall lower number of cycles for Im2col. However, note that the gathering overhead is much higher for layers with a larger $C_{in}$ (see Eq.~\ref{eq:gather}), so the ranking among the two depends on the layer topology. Further, the Indirect kernel benefits from a nearly null additional memory, often improving the performance when considering the effect of tiling on large layers. These aspects will be better discussed in Section \ref{subsec:use_cases}.

\subsection{Comparison with State-of-the-art NN backends}
\label{subsec:SoaCompare}
\begin{figure}[t]
  \centering
\includegraphics[width=.81\columnwidth]{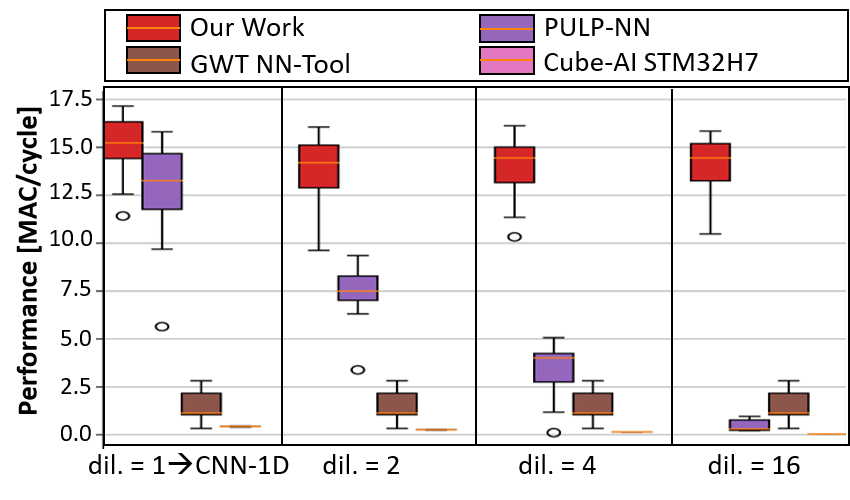}
  \vspace{-0.5cm}
  \caption{\revDAC{Comparison with state-of-the-art CNN backends for edge devices.}}\label{fig:comparison}
  \vspace{-0.6cm}
\end{figure}

\revDAC{Figure~\ref{fig:comparison} shows a complete comparison between our toolkit and state-of-the-art backends. The figure reports the performance (in MAC/cycle) for layers with dilation $d \in (1,2,4,16)$. For each value of $d$, the box plots aggregate the results of multiple layers with different shapes. We used $T \in (16, 64)$, $K \in (3, 5, 7)$, and $C_{in} =  C_{out} \in (32, 64, 128, 256)$. }
%

\revDAC{Our toolkit consistently outperforms the state-of-the-art across different layer shapes and dilation values.
In particular, we observe a dramatically higher performance than GWT NN-Tool, i.e., $9.7\times$ on average.} This is due to the CHW format used in its convolutions, which converts to a strongly sub-optimal CT layout for 1D kernels (see Sec. \ref{subsec:opt} for details).
With respect to PULP-NN, we observe a slightly higher performance for $d=1$ \revDAC{($1.2\times$)}, thanks to the elimination of unnecessary im2col buffers and to the optimization of the internal MatMul loop execution for 1D data.
The benefit increases significantly for larger dilation factors \revDAC{(e.g. $28.9\times$ for $d = 16$)} since, as mentioned, PULP-NN kernels do not support this fundamental 1D-convolution propriety, which has to be reproduced interleaving weights with 0s, increasing the filter dimension from $K$ to $(K-1)\times d + 1$ and inserting a large number of useless MAC operations.
%
%
\revDAC{With respect to Cube-AI, we obtain a speed-up between 34.7$\times$ (for $d=1$) to 354$\times$ (for $d=16$). The higher speedup for higher $d$ has the same motivation discussed for PULP-NN.
\revISPLED{Comparing both of them during single-core execution, our toolkit still demonstrates 4.7$\times$, 7.0$\times$, 13.0$\times$, and 47.6$\times$ higher MACs/cycle.}
Considering the energy efficiency in GMAC/s/W, the  improvement over PULP-NN and GWT NN-Tool is proportional to the speedup, given that the execution platform is the same. Compared to Cube-AI, instead, considering the best energy configuration for both STM32H7 and GAP8, we obtain 33.1$\times$, 50.0$\times$, 92.3$\times$ and 338.4$\times$ higher efficiency on average for $d$ = 1, 2, 4, 16.
Importantly, notice that $d=1$ corresponds to a standard 1D CNN layer; hence, the first set of box plots show that our tool is outperforming the state-of-the-art not just on dilated TCNs, but also on classical 1D-CNNs.
}

\subsection{Complete use cases}
\label{subsec:use_cases}
%
\begin{table*}[]
\centering
\caption{End-to-end comparison on three TCNs architectures for different tasks. Abbreviations: OOM: Out of Memory.}
\vspace{-0.2cm}
\label{tab:UseCases}
\begin{tabular}{l|cccc|c|c|cc}
\hline
                            & \multicolumn{4}{c|}{Our Work}                                                                                              & \multirow{2}{*}{PULP-NN} & \multirow{2}{*}{NN-Tool} & \multicolumn{2}{c}{\multirow{2}{*}{Cube-AI}} \\ 
                            & \multicolumn{1}{l}{Indirect} & \multicolumn{1}{l}{No-Im2col} & \multicolumn{1}{l}{Im2col} & \multicolumn{1}{l|}{Optimizer} &                          &                          & \multicolumn{2}{c}{}                          \\ \hline
Platforms [MCU]                   & \multicolumn{6}{c|}{GAP8, 1xRISC-V + 8xRISC-V}    & STM32H7, 1xCortexM7                 & STM32L4, 1xCortexM4              \\ 
Power [mW] /   Freq. [MHz]  & \multicolumn{6}{c|}{51 mW / 100 MHz}       & 234 mW / 480 MHz        & 10 mW / 80MHz        \\ \hline
\multicolumn{9}{l}{\textbf{TEMPONet -   Gesture Recognition - Parameters: 86.5k - MACs: 15.1M}}                                                                                                                                                                 \\ \hline
Time/Inference   [ms]       & 17.18                         & 29.79                          & 13.94                       & 13.60                         & 38.78                    & 103.10                   & 138.29                  & 1408.16              \\ 
Energy[mJ]                  & 0.88                          & 1.52                           & 0.71                        & 0.69                          & 1.98                     & 5.26                     & 32.36                   & 14.08                \\ 
MACs/cycle                  & 8.80                          & 5.07                           & 10.84                       & 11.11                         & 3.90                     & 1.47                     & 0.23                    & 0.13                 \\ 
Throughput   [GMAC/s]       & 0.88                          & 0.51                           & 1.08                        & 1.11                          & 0.39                     & 0.15                     & 0.11                    & 0.01                 \\ 
En.Efficiency   [GMACs/s/W] & 17.25                         & 9.95                           & 21.26                       & 21.79                         & 7.64                     & 2.87                     & 0.47                    & 1.07                 \\ \hline
\multicolumn{9}{l}{\textbf{ResTCN -   Sound Generation - Parameters: 1.13M - MACs: 18.1M}}                                                                                                                                                                      \\ \hline
Time/Inference   [ms]       & 23.91                         & OOM                           & 24.66                       & 23.45                         & OOM                     & 568.25                   & 215.79                  & OOM                 \\ 
Energy[mJ]                  & 1.22                          & OOM                           & 1.26                        & 1.20                          & OOM                     & 28.98                    & 50.49                   & OOM                 \\ 
MACs/cycle                  & 7.57                          & OOM                           & 7.34                        & 7.72                          & OOM                     & 0.32                     & 0.17                    & OOM                 \\ 
Throughput   [GMAC/s]       & 0.76                          & OOM                           & 0.73                        & 0.77                          & OOM                     & 0.03                     & 0.08                    & OOM                 \\ 
En.Efficiency   [GMACs/s/W] & 14.84                         & OOM                           & 14.39                       & 15.13                         & OOM                     & 0.62                     & 0.36                    & OOM                 \\ \hline
\multicolumn{9}{l}{\textbf{ResTCN -   Language Modeling - Parameters: 2.7M - MACs: 135M}}                                                                                                                                                                      \\ \hline
Time/Inference   [ms]       & 171.00                        & OOM                           & 665.91                      & 168.51                        & OOM                     & 4490.00                  & 2315.25                 & OOM                 \\ 
Energy[mJ]                  & 8.72                          & OOM                           & 33.96                       & 8.59                          & OOM                     & 228.99                   & 541.77                  & OOM                 \\ 
MACs/cycle                  & 7.89                          & OOM                           & 2.03                        & 8.01                          & OOM                     & 0.30                     & 0.12                    & OOM                 \\ 
Throughput   [GMAC/s]       & 0.79                          & OOM                           & 0.20                        & 0.80                          & OOM                     & 0.03                     & 0.06                    & OOM                 \\ 
En.Efficiency   [GMACs/s/W] & 15.48                         & OOM                           & 3.98                        & 15.71                         & OOM                     & 0.59                     & 0.25                    & OOM                 \\ \hline
\end{tabular}
\vspace{-0.7cm}
\end{table*}
In this section, we use our toolkit and the comparison baselines to implement three complete TCN architectures of different size.
\revLUCA{Table~\ref{tab:UseCases} reports a complete comparison on three networks, TEMPONet~\cite{zanghieri2019robust} for gesture recognition, and two ResTCNs from~\cite{bai2018empirical}, for sound generation and language modeling, respectively. }
While the number of layers of the 3 networks is similar (9, 8, and 10), the number of filters per layer, hence the number of parameters and MACs, is increasingly high.
Specifically, TEMPONet has a modular structure that progressively shrinks the time dimension while increasing the number of channels up to 128~\cite{zanghieri2019robust}, while the other two TCNs maintain a constant $T$ (16 and 50) with respectively 150 and 450 channels per layer.

We draw two main observations from these experiments.
First, integrating our kernel models and kernel selection in the tiling optimizer leads to up to $4.0\times$ speed-up compared to always using a single kernel implementation.
While mapping all the layers of TEMPONet to the Im2col kernel leads to a near-optimal implementation, the same strategy applied to the language-modeling TCN yields 4$\times$ lower performance than a per-layer selection.
Similarly, considering solely the Indirect kernel results in $1.3\times$ lower performance on TEMPONet.
%
%
Therefore, choosing the appropriate kernel for each layer is key to maximize performance. In general, for layers with $d=1$, No-im2col reaches the highest performance, while Im2col and Indirect are optimal for layers with $d > 1$ with a low/high number of channels, respectively.

We also compare our best performance with state-of-the-art full-stack tools, including their back-end and, when available, their tiling mechanism.
We see a minimum speed-up of $2.9\times$ compared to the DORY+PULP-NN stack for the 3 networks. However, 2 out of 3 cannot be implemented using this tool given the high memory overhead of the Im2col support buffer, which does not fit L1 memory, preventing any form of tiling (i.e., even a single convolution cannot be computed entirely from L1 data). 
Directly storing the whole networks in the slow GAP-8 L2 memory (512KB) leads to a slowdown of more than 4$\times$.
When we compare to Cube-AI and GWT NN-Tool on all the networks, we observe speed-ups of \revDAC{$7.6\times$ to $103\times$, with at least $20.3\times$} lower energy.
With respect to the original TEMPONet deployment of~\cite{zanghieri2019robust}, which uses a customized version of PULP-NN, we achieve 11.53 vs. 7.73 MAC/cycles, i.e. 49\% better performance. This corresponds to an end-to-end latency of 13.6ms per inference, i.e. below the real-time constraint of 15ms indicated in~\cite{zanghieri2019robust}, obtained consuming 0.69mJ, with GAP-8 running at 100MHz \revLUCA{and 51 mW, compatible with battery powered edge applications}. In the same conditions, an inference with the other two TCNs completes in 23.45 / 168.5ms and consumes 1.2 / 8.59mJ for sound/language processing respectively.
Noteworthy, the ResTCN for both sound generation and language modeling does not fit the memory constraint nor on STM32L4 neither in GAP8 when PULP-NN baseline. 
\section{Conclusion}
\label{sec:conclusion}
The efficient processing of time-series is a key element for many DL-based IoT applications.
To fill the gap between new NN topologies and their efficient execution, we have proposed a software toolkit to optimize the performance of TCNs on smart edge-nodes.
We have shown that thanks to multiple kernel implementations, we can maintain top performance on different real-world networks for various use-cases, reaching $3\times$ to $103\times$ speed-up and up to $63.0\times$ higher energy-efficiency compared to other software back-ends on multi- and single- core edge-nodes.


\footnotesize
\bibliographystyle{IEEEtran}
\tiny

\end{document}